\documentclass[10pt,twocolumn,letterpaper]{article}

\usepackage[pagenumbers]{cvpr} 

\usepackage{graphicx}
\usepackage{amsmath}
\usepackage{amssymb}
\usepackage{booktabs}
\usepackage{anyfontsize}

\usepackage[pagebackref,breaklinks,colorlinks]{hyperref}

\usepackage[capitalize]{cleveref}
\usepackage{import}
\usepackage{arydshln}
\usepackage{multirow}
\usepackage{lipsum}
\usepackage{inconsolata} 
\usepackage{amssymb}
\usepackage{pifont}
\usepackage{clock}
\usepackage[clock]{ifsym}
\usepackage{float}
\usepackage{graphicx}
\usepackage{amsmath}
\usepackage{soul}
\usepackage[table]{xcolor}
\usepackage{enumerate}
\usepackage{enumitem}
\usepackage{subcaption}
\usepackage{adjustbox}

\usepackage{algorithm}
\usepackage{algorithmic}
\definecolor{Gray}{gray}{0.9}
\def\cvprPaperID{7945} 
\def\confName{CVPR}
\def\confYear{2023}

\begin{document}

\title{SuperDisco: Super-Class Discovery Improves
Visual Recognition for the Long-Tail}


\author{
Yingjun Du\textsuperscript{1},
Jiayi Shen\textsuperscript{1},
Xiantong Zhen\textsuperscript{1,2}\thanks{Currently with United Imaging Healthcare, Co., Ltd., China.}~,
Cees G. M. Snoek\textsuperscript{1} \\
\textsuperscript{1}AIM Lab, University of Amsterdam \textsuperscript{2}Inception Institute of Artificial Intelligence}

\maketitle

\begin{abstract}
Modern image classifiers perform well on populated classes, while degrading considerably on tail classes with only a few instances. Humans, by contrast, effortlessly handle the long-tailed recognition challenge, since they can learn the tail representation based on different levels of semantic abstraction, making the learned tail features more discriminative. This phenomenon motivated us to propose SuperDisco, an algorithm that discovers super-class representations for long-tailed recognition using a graph model. We learn to construct the super-class graph to guide the representation learning to deal with long-tailed distributions. Through message passing on the super-class graph, image representations are rectified and refined by attending to the most relevant entities based on the semantic similarity among their super-classes. Moreover, we propose to meta-learn the super-class graph under the supervision of a prototype graph constructed from a small amount of imbalanced data. By doing so, we obtain a more robust super-class graph that further improves the long-tailed recognition performance. The consistent state-of-the-art experiments on the long-tailed CIFAR-100, ImageNet, Places and iNaturalist demonstrate the benefit of the discovered  super-class graph for dealing with long-tailed distributions.
\end{abstract}

\section{Introduction}
This paper strives for long-tailed visual recognition. A computer vision challenge that has received renewed attention in the context of representation learning, as real-world deployment demands moving from balanced to imbalanced scenarios. Three active strands of work involve class re-balancing \cite{estabrooks2004multiple,han2005borderline,liu2008exploratory,kang2019decoupling,wang2020devil}, information augmentation \cite{shu2019meta, ren2018learning, kim2020m2m} and module improvement \cite{huang2016learning, zhang2017range, kang2021exploring}. Each of these strands is intuitive and has proven empirically successful. However, all these approaches seek to improve the classiﬁcation performance of the original feature space. In this paper, we instead explore a graph learning algorithm to discover the imbalanced super-class space hidden in the original feature representation.

The fundamental problem in long-tailed recognition~\cite{liu2019large,zhang2021deep,kang2019decoupling,garcin2021pl} is that the head features and the tail features are indistinguishable. Since the head data dominate the feature distribution, they cause the tail features to fall within the head feature space. Nonetheless, humans effortlessly handle long-tailed recognition  \cite{anderson2007long, ferguson2014big} by leveraging semantic abstractions existing in language to gain better representations of tail objects. This intuition hints that we may discover the semantic hierarchy from the original feature space and use it for better representations of tail objects.
\begin{figure} [t]
\centering
   \begin{minipage}{.49\columnwidth}
    \subfloat[ 100 original classes]{
      \includegraphics[width=.85\columnwidth]{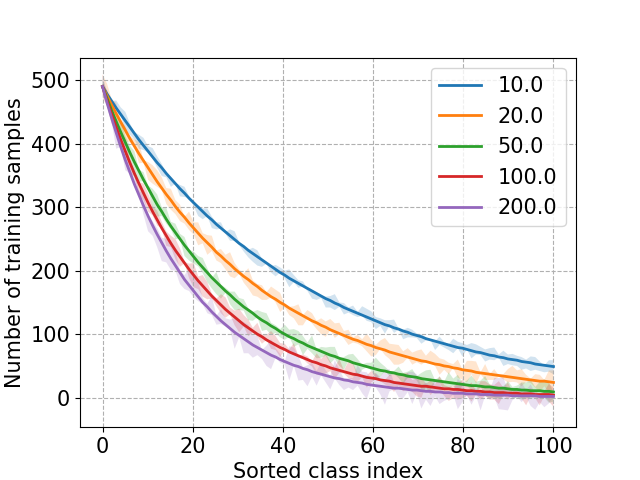}}
\end{minipage}
   \begin{minipage}{.49\columnwidth}
   \subfloat[20 ground truth super-classes]{
      \includegraphics[width=.85\columnwidth]{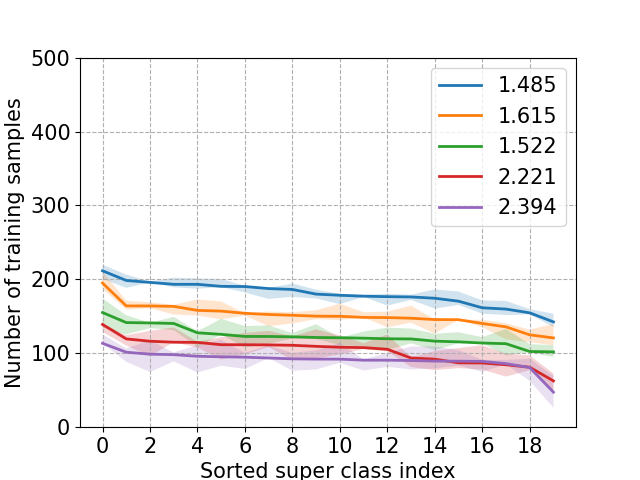}}
\end{minipage}
   \begin{minipage}{.49\columnwidth}
   \subfloat[16 discovered super-classes]{
      \includegraphics[width=.85\columnwidth]{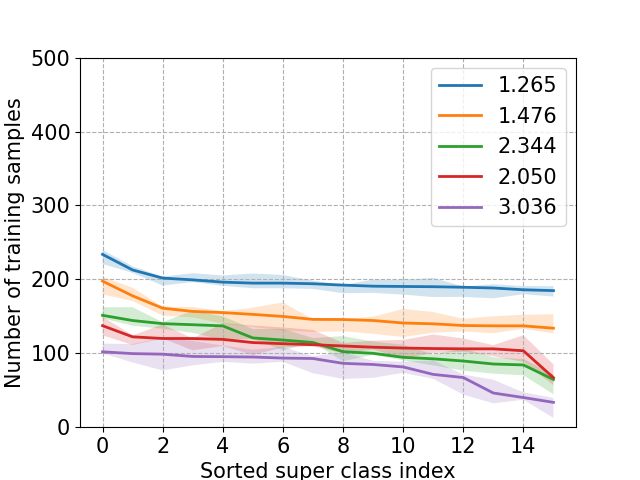}}
\end{minipage}
   \begin{minipage}{.49\columnwidth}
   \subfloat[32 discovered super-classes]{
      \includegraphics[width=.85\columnwidth]{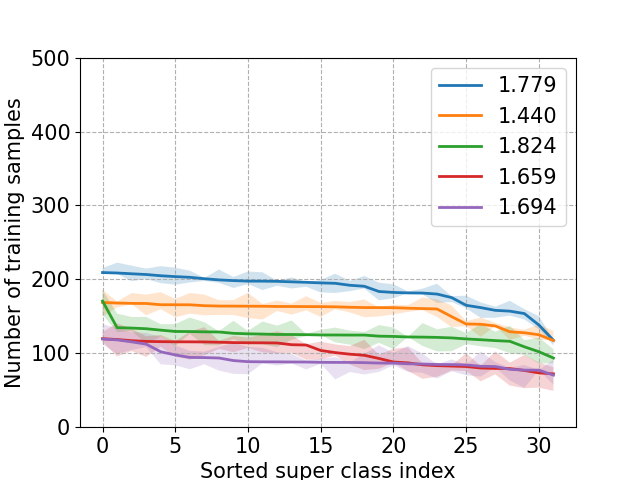}}
\end{minipage}
\caption{\textbf{SuperDisco learns to project the original class space (a) into a relatively balanced super-class space.} Different color curves indicate the different imbalance factors on the long-tailed CIFAR-100 dataset. Like the 20 super-class ground truth (b) our discovered super-classes for 16 super-classes (c) or 32 super-classes (d) provide a much better balance than the original classes.}
	\label{fig:distribution}
\end{figure}
Moreover, intermediate concepts have been shown advantageous for classiﬁcation \cite{chen2020concept, koh2020concept} by allowing the transfer of shared features across classes. Nevertheless, it remains unexplored to exploit intermediate super-classes in long-tailed visual recognition that  rectify and refine the original features.

In the real world, each category has a corresponding super-class, \eg, \textit{bus}, \textit{taxi}, and \textit{train} all belong to the \textit{vehicle} super-class. This observation raises the question: \emph{are super-classes of categories also distributed along a long-tail?} We find empirical evidence that within the super-class space of popular datasets, the long-tailed distribution almost disappears, and each super-class has essentially the same number of samples. 
In Figure~\ref{fig:distribution}, we show the number of training samples for each of the original classes and their corresponding super-classes in the long-tailed CIFAR-100 dataset.  We observe the data imbalance of super-classes is considerably lower than those of the original classes.  This reflects the fact that the original imbalanced data hardly affects the degree of imbalance of the super-classes, which means the distribution of the super-classes and original data is relatively independent. These balanced super-class features could be used to guide the original tail data away from the dominant role of the head data, thus making the tail data more discriminative. Therefore, if the super-classes on different levels of semantic abstraction over the original classes can be accurately discovered, it will help the model generalize over the tail classes. As not all datasets provide labels for super-classes, we propose to learn to discover the super-classes in this paper.

Inspired by the above observation, we make in this paper two algorithmic contributions. First, we propose in Section~\ref{sec:hcg} an algorithm that learns to discover the super-class graph for long-tailed visual recognition, which we call SuperDisco. We construct a learnable graph that discovers the super-class in a hierarchy of semantic abstraction to guide feature representation learning. By message passing
on the super-class graph, the original features are rectified and refined, which attend to the most relevant entities according to the similarity between the original image features and super-classes.  
Thus, the model is endowed with the ability to free the original tail features from the dominance of the head features using the discovered and relatively balanced super-class representations. Even when faced with the severe class imbalance challenges, 
\eg, iNaturalist, our SuperDisco can still refine the original features by finding a more balanced super-class space using a more complex hierarchy.
As a second contribution, we propose in Section~\ref{sec:mhcg} a meta-learning variant of our SuperDisco algorithm to discover the super-class graph, enabling the model to achieve even more balanced image representations. To do so, we use a small amount of balanced data to construct a prototype-based relational graph, which captures the underlying relationship behind samples and alleviates the potential effects of abnormal samples. 
Last, in Section~\ref{sec:exp} we report experiments on four long-tailed benchmarks: CIFAR-100-LT, ImageNet-LT, Places-LT, and iNaturalist, and verify that our discovered super-class graph performs better for tail data in each dataset.
Before detailing our contributions, we first embed our proposal in related work.

\section{Related work}

\textbf{Long-tailed recognition.}
Several strategies have been proposed to address class imbalance in recognition.  We categorize them into three groups. Those in the first group are based on class re-balancing~\cite{cui2019class,jamal2020rethinking,liu2019large,zhang2021distribution}, which balance the training sample numbers of different classes during model training. Class re-balancing methods also could be categorized into three different groups, \ie, re-sampling~\cite{han2005borderline, liu2008exploratory, kang2019decoupling, wang2020deep}, cost-sensitive learning~\cite{elkan2001foundations, zhou2005training, sun2007cost, zhao2018adaptive, zhang2018online, li2022equalized} and logit adjustment~\cite{menon2020long, hong2020disentangling, tian2020posterior, tang2020long}. Class re-balancing methods improve the performance of the tail classes at the expense of the performance of the head classes.
The second group is based on information augmentation, introducing additional information into model training to improve long-tailed learning performance. We identify four kinds of methods in the information augmentation scope, \ie, transfer learning, which includes head-to-tail knowledge transfer~\cite{yin2019feature, liu2020deep, chu2020feature, wang2021rsg}, knowledge distillation~\cite{li2021self, hu2020learning, xiang2020learning}, model pre-training~\cite{cui2018large, yang2020rethinking, karthik2021learning} and self-training~\cite{he2021re, wei2021crest, zhang2021mosaicos}.  The third group focuses on improving network modules in long-tailed learning. This group includes representation learning~\cite{zhang2017range, ouyang2016factors, dong2017class}, classifier learning~\cite{kang2019decoupling, yin2019feature, liu2020deep, wu2021adversarial, liu2021gistnet}, decoupled training~\cite{kang2019decoupling, kang2021exploring, zhong2021improving}, and ensemble learning~\cite{zhou2020bbn, guo2021long}. These methods introduce additional computation costs for increased performance. 
Our method belongs to the third group as it aims to learn a better representation  of unbalanced training samples by the super-class graph, which is unexplored for long-tail recognition. 

\textbf{Super-class learning.} Super-class learning adds super-class labels as intermediate supervision into traditional deep learning.  A super-class guided network~\cite{li2021sgnet}  integrated the high-level semantic information into the network for image classification and object detection, which took two-level class annotations that contain both super-class and finer class labels. In \cite{dehkordi2022multi}, a two-phase multi-expert architecture was proposed for still image action recognition, which includes fine-grained and coarse-grained phases. However, they leveraged the ground truth of the super-class as supervision during the coarse-grained phase. Wu \etal~\cite{wu2020solving} propose a taxonomic classifier to address the long-tail recognition problem, which classified each sample to the level that the classifier is competent. Zhou \etal~\cite{zhou2018deep} clustered the original categories into super-classes to produce a relatively balanced distribution in the super-class space, which also leveraged the ground truth of the super-class in the training phase.
In contrast with the previous super-class learning, we do not use ground truth to group the original categories into the super-class space.
To the best of our knowledge, no work exists that relies on graph learning to discover the super-class for long-tailed visual recognition, thus motivating this work.

\begin{figure*} [t!]
\centering
\includegraphics[width=0.95\linewidth]{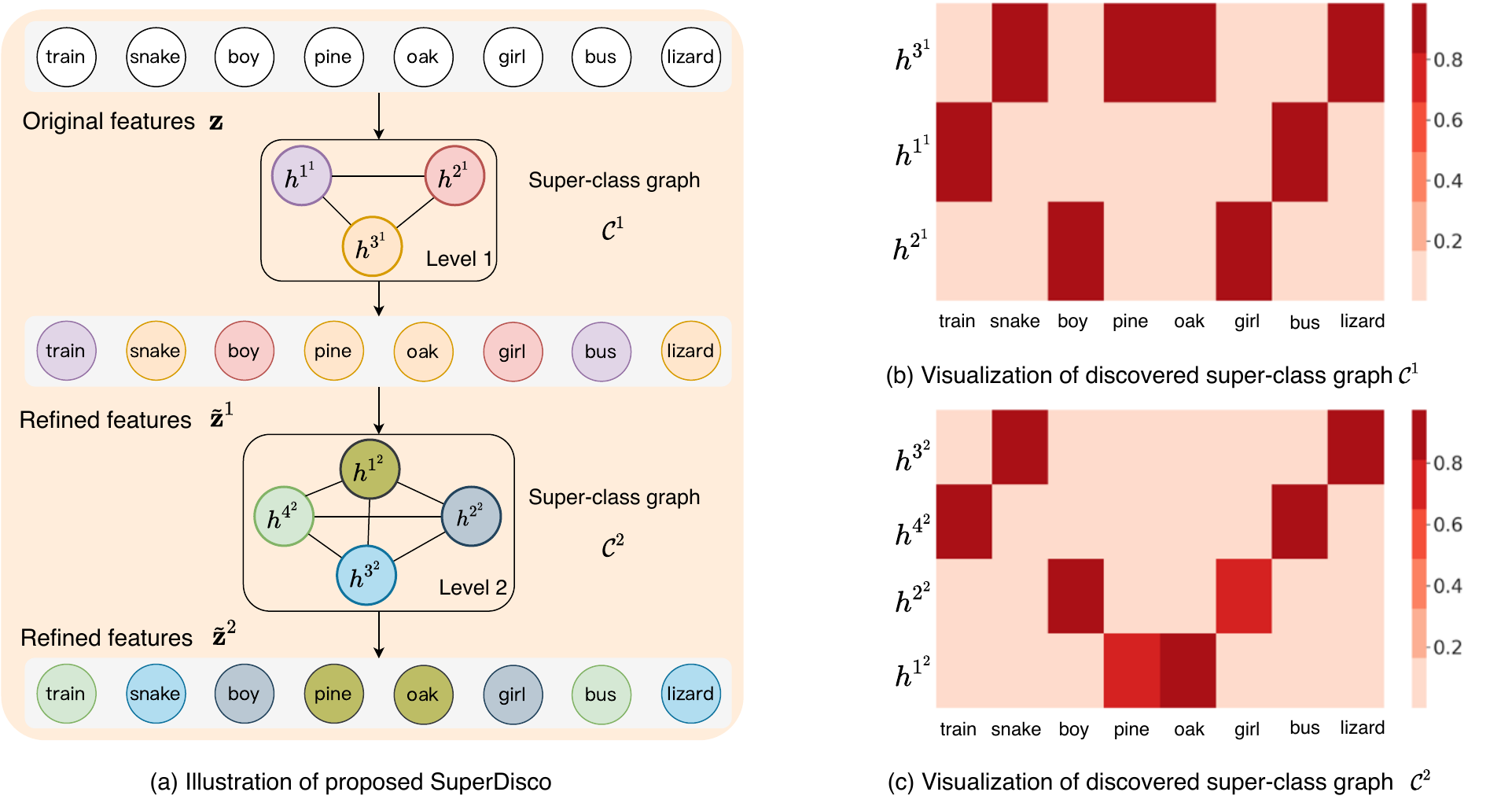}
\caption{\textbf{Illustration of proposed SuperDisco} (a) and visualization of the similarity between the classes and  discovered super-class at different levels (b), (c). In (a), we show  two levels of super-class graphs $\mathcal{C}^1$ and $\mathcal{C}^2$ . The colour in  each graph represents the discovered super-class. SuperDisco discovers the potential super-class at different levels hidden in  each category from (b) and (c). 
$\mathcal{C}^1$  roughly categorizes the original classes into three relatively balanced super-classes, and  $\mathcal{C}^2$  then  finely categorize them into four more balanced  super-classes.}
	\label{fig:framework}
\end{figure*}

\textbf{Graph neural networks.}  Recently, several graph neural network models (GNN) have been proposed to exploit the structures underlying graphs to benefit a variety of applications. There are two main research lines of GNN methods: non-spectral methods and spectral methods.  The spectral methods~\cite{defferrard2016convolutional, kipf2016semi, henaff2015deep, bruna2013spectral} focus on learning graph representations in a spectral domain, in which the learned filters are based on Laplacian matrices. 
The non-spectral methods~\cite{velivckovic2017graph, hamilton2017inductive} develop an aggregator to aggregate a local set of features. 
Note that, message passing~\cite{gilmer2017neural} is a key mechanism that allows GCNs and other graph neural networks to capture complex relationships and dependencies between nodes in a graph, and is a major reason why they have been successful in a variety of tasks involving graph-structured data.
Our method belongs to the non-spectral methods, which leverage a GNN as the base architecture to discover the super-class representation. Our proposed super-class graph would refine and rectify the original imbalanced feature to a relatively balanced feature space, which has  not been explored for long-tail recognition either.

\textbf{Meta-learning for the long-tail.} Meta-learning or learning to learn \cite{schmidhuber1987evolutionary,bengio90,thrun_metalearning, zhen2020icml, zhen2020learning}, is a learning paradigm where a model is trained on the distribution of tasks so as to enable rapid learning on new tasks.  Ren \etal~\cite{ren2018learning} first proposed meta-learning for the long-tailed problem by reweighting training examples.  Shu \etal~\cite{shu2019meta} proposed Meta-weight-Net to adaptively extract sample weights to guarantee robust recognition in the presence of training data bias. Li \etal~\cite{li2021metasaug} introduced meta-semantic augmentation for long-tailed recognition, which produces diversified augmented samples by translating features along many semantically meaningful directions by meta-learning. 
Our uniqueness is that our model aims to discover an improved super-class representation by meta-learning, which enables the original feature representation to adjust its corresponding higher-level super-class space.

\section{Learning to discover the super-class graph}
\label{sec:hcg}
In this section, we discuss how to learn to discover the super-class  graph from the training samples and then expand on how to leverage such a graph to benefit the unbalanced data by refining the feature representations of samples. The overall illustration of  SuperDisco and a visualization of the discovered super-class hierarchy are shown in Figure~\ref{fig:framework}. 

\textbf{Preliminary.} For long-tailed visual recognition, the goal is to learn an image classification model from an imbalanced training set and to evaluate the model on a balanced test set. We first define the notation for long-tailed recognition used throughout our paper. We define a training input $x_k \in \mathbb{R}, i \in \{{1, \cdots, n}\}$, where $n$ is the number of training samples in the dataset. The corresponding labels are $y_k \in {1, 2, \cdots, C}$, where $C$ is the number of classes. 
Let $n_j$ denote the number of training samples for the class $j$. Here, we assume that $n_i \geq n_j$ when $i < j$ shows the long-tailed problem simply. In this work, we typically consider a deep network model with three main components:  a feature extractor $f(\cdot)$, a proposed graph model $g(\cdot)$ and a classifier $h(\cdot)$. The feature extractor $f(\cdot)$ first extracts an image representation as $\mathbf{z} {=} f (x;\theta)$, which is then fed into the proposed graph model to refine a new representation as  $g(\mathbf{z}; \phi) {=} \tilde{\mathbf{z}}$. The final class prediction $\tilde{y}$ is given by a classifier function $h(\cdot)$, \ie, $\tilde{y} {=} \arg\max h(\tilde{\mathbf{z}}; \psi)$. Before detailing our approach, we add Table~\ref{tab:symbol} to detail  the meaning of each symbol for easy lookup.

\textbf{SuperDisco.}
We construct the super-class graph to organize and distill knowledge from the training process. The vertices represent different types of super-classes (\eg, the common contour between \textit{birds} and \textit{airplanes}) and the edges are automatically constructed to reflect the relationship between different super-classes. Our super-class graph contains multiple levels, which is closer to the relationship between various objects in the real world. 
Before detailing the structure, we first explicate why the multi-level super-class graphs are preferred over a flat super-class graph: a single level of  super-class  groups is likely insufﬁcient to model complex task relationships in real-world applications; for example, the similarities among different bird species are high, but there are also similarities between birds and mammals, \eg, they are both animals.

We assume the vertex representation \begin{small}$g$\end{small} as \begin{small}$\mathbf{h}^g\in \mathbb{R}^{d}$\end{small}, and define the super-class graph as \begin{small}$\mathcal{C}^l{=}(\mathbf{H}^l_{\mathcal{C}}, \mathbf{A}^l_{\mathcal{C}})$\end{small}, where \begin{small}$\mathbf{H}^l_{\mathcal{C}}{=}\{\mathbf{h}^{i^l}|\forall i^l \in [1,C^l]\}\in \mathbb{R}^{C^l\times d}$\end{small} is the vertex feature matrix of the $l$-th super-class  level  and \begin{small}$\mathbf{A}^l_{\mathcal{C}}{=}\{A^l_{\mathcal{C}}(\mathbf{h}^{i^l},\mathbf{h}^{j^l})|\forall i^l,j^l\in[1,C^l]\}\in \mathbb{R}^{C^l\times C^l}$\end{small} is the vertex adjacency matrix in the $l$-th super-class level, $C^l$ denotes the number of vertices in the $l$-th super-class  level. 
Our vertex representation  $\mathbf{H}^l_{\mathcal{C}}$ of the super-class  graph is defined to get parameterized and learned during training.  The initial vertex representations of each super-class  level are randomly initialized, which encourages diversity of the discovered super-classes. 

Next, we introduce how to compute the edge weight $A_{\mathcal{C}}^l$ in the super-class  graph. The edge weight $A_{\mathcal{C}}^l(\mathbf{h}^{i^l},\mathbf{h}^{j^l})$ between a pair of vertices \begin{small}$i$\end{small} and \begin{small}$j$\end{small} is gauged by the similarity between them.
Formally:
\begin{equation}
    A^l_{\mathcal{C}}(\mathbf{h}^{i^l},\mathbf{h}^{j^l}){=}\sigma(\mathbf{W}_c^l(|\mathbf{h}^{i^l}-\mathbf{h}^{j^l}|/\gamma^l_c)+\mathbf{b}^l_c),
\label{eq:hc_e}
\end{equation}
where \begin{small}$\mathbf{W}_c^l$\end{small} and \begin{small}$\mathbf{b}^l_c$\end{small} indicate learnable parameters of the $l$-th super-class  level,  \begin{small}$\gamma_c^l$\end{small} of $l$-th super-class  level is a scalar and \begin{small}$\sigma$\end{small} indicates the Sigmoid function, which normalizes the weight between $0$ and $1$. 
To adjust the representation of training samples by  the involvement of super-classes, we first query the training samples in the super-class  graph to obtain the relevant super-class. In light of this, we construct a new graph $\mathcal{R}$, which adds the original sample feature $\mathbf{z}$ to the super-class  graph. We define $\mathbf{z}^l$ as the refined feature after the $l$-th super-class  graph. Here we define graph \begin{small}$\mathcal{R}^l{=}(\mathbf{H}^l_{\mathcal{R}}, \mathbf{A}^l_{\mathcal{R}})$\end{small},   where \begin{small}$\mathbf{H}^l_{\mathcal{R}}{=}\{[\mathbf{z}^l, \mathbf{h}^{i^l}]|\forall i^l \in [1,C^l]\}\in \mathbb{R}^{(C^l + 1)\times d}$\end{small} denotes the vertex feature matrix of the $l$-th super-class  level,  and \begin{small}$\mathbf{A}^l_{\mathcal{R}}{=}\{[A^l_{\mathcal{R}}(\mathbf{h}^{i^l}, \mathbf{z^l}), 
A^l_{\mathcal{R}}(\mathbf{h}^{i^l},\mathbf{h}^{j^l})]|\forall i^l,j^l\in[1,C^l]\}\in \mathbb{R}^{C^{l+1}\times C^{l+1}}$\end{small} denotes the vertex adjacency matrix in the $l$-th super-class  level. The link between $\mathbf{z}^l$ and vertex $\mathbf{h}^i$ in the hierarchical graph is constructed by their similarity. In particular, analogous to the definition of weight in the super-class  graph in Eq.~(\ref{eq:hc_e}), the weight 
$\begin{small}A^l_{\mathcal{R}}(\mathbf{h}^{i^l}, \mathbf{z^l})\end{small}$ is constructed as:
\begin{equation}
    A^l_{\mathcal{R}}(\mathbf{h}^{i^l},\mathbf{z^l}){=}\sigma(\mathbf{W}_r^l(|\mathbf{h}^{i^l}-\mathbf{z^l}|/\gamma^l_r)+\mathbf{b}^l_r),
\label{eq:hc_r}
\end{equation}
where \begin{small}$\mathbf{W}_r^l$\end{small} and \begin{small}$\mathbf{b}^r_c$\end{small} indicate learnable parameters of the $l$-th super-class  level,  \begin{small}$\gamma_r^l$\end{small} of the $l$-th super-class  level is a scalar.


\begin{table}[t]
    \centering
    \scalebox{0.95}{\begin{tabular}{l l}
       \hline
Notation & Description \\
\hline
$\textbf{h}$ & Vertex representation \\
$\mathbf{z}$ & Original sample feature \\ 
$\mathcal{C}$ & Super-class graph \\
$\mathbf{H}_{\mathcal{C}}$ & Vertex feature matrix of $\mathcal{C}^l$ \\
$\mathbf{A}_{\mathcal{C}}$ & Vertex adjacency matrix of $\mathcal{C}^l$ \\
$\mathcal{R}$  & Graph which adds $\mathbf{z}$ to graph $\mathcal{C}^l$ \\
$\mathcal{P}$ & Prototype graph \\
$\mathbf{C}_\mathcal{P}$ &Vertex feature matrix of $\mathcal{P}$ \\
$\mathbf{A}_\mathcal{P}$ &  Vertex adjacency matrix of  $\mathcal{P}$  \\
$\mathcal{S}$ & Super graph which connecting $\mathcal{P}$ and $\mathcal{C}$ \\
$\mathbf{A}$ & Vertex feature matrix  of $\mathcal{S}$ \\
$\mathbf{M}$ & Vertex adjacency matrix of $\mathcal{S}$ \\
\hline
    \end{tabular}}
    \caption{Summary of the core notation used for SuperDisco.}
    \label{tab:symbol}
    \vspace{-2mm}
\end{table}

After constructing the new graph $\mathcal{R}$, we propagate the most relevant super-class  by message passing~\cite{gilmer2017neural} from the discovered super-classes  $\mathcal{C}$ to the features $\mathbf{z}^l$ by introducing a Graph Neural Network (GNN). The message passing operation over the graph is formulated as:
\begin{equation}
    \mathbf{H}^{(m+1)}_{\mathcal{R}}{=}\texttt{MP}(\mathbf{A}^l_{\mathcal{R}},\mathbf{H}^{(m)}_{\mathcal{R}};\mathbf{W}^{(m)}),
    \label{eq:mpc}
\end{equation}
where \begin{small}$\texttt{MP}(\cdot)$\end{small} is the message passing function, \begin{small}$\mathbf{H}^{(m)}$\end{small} is the vertex embedding after $m$ layers of GNN and \begin{small}$\mathbf{W}^{(m)}$\end{small} is a learnable weight matrix of layer \begin{small}$m$\end{small}.  After stacking \begin{small}$M$\end{small} GNN layers, we get the information-propagated feature representation $\tilde{\mathbf{z}}^L$ for each level of the super-class  graph $\mathcal{C}$. 
Once we obtain the refined representation $\tilde{\mathbf{z}}^L$ for a training sample by the super-class  graph, we feed them into the classifier to make the predictions and compute the corresponding loss, \ie, Cross-entropy loss for optimization. Using gradient descent, we then update the super-class graph $\mathcal{C}$. To be able to discover a more accurate super-class graph in the face of severe imbalance problems, we propose meta-learning super-class graph discovery in the next section.

\section{Meta-learning super-class graph discovery}
\label{sec:mhcg}
To explore and exploit a more accurate and richer super-class  graph, we propose the Meta-SuperDisco to discover the super-class graph using meta-learning, making the model more robust.
In the traditional meta-learning setting~\cite{finn2017model, ravi2016optimization}, it includes meta-training tasks and meta-test task. Each task contains a support set $\mathcal{S}$ and a query set $\mathcal{Q}$.
Each task is first trained by  $\mathcal{S}$ to get the task-specific learner and $\mathcal{Q}$ optimizes this learner to update the meta-learner.  
For long-tailed recognition with meta-learning, previous works~\cite{shu2019meta, ren2018learning} randomly sample a small amount of balanced  data denoted as $\mathcal{M}$. The imbalanced data and the small balanced  data  can be seen as  $\mathcal{S}$ and $\mathcal{Q}$ in the training phase. The goal of meta-learning for long-tailed recognition is to use a small set of balanced data to optimize the model obtained from unbalanced data.  We follow~\cite{ren2018learning, shu2019meta} by randomly selecting the same number of samples (\eg, ten) as $\mathcal{M}$ per class from the training set.

\textbf{Meta-SuperDisco.}
To meta-learn the super-class graph, we construct a prototype graph $\mathcal{P}$ from   $\mathcal{M}$, since  $\mathcal{M}$ is a balanced dataset.  The prototype graph extracts the sample-level relation information, which captures the underlying relationship behind samples and alleviates the potential effects of abnormal samples. For the prototype graph, we need to compute the prototype of each category~\cite{snell2017prototypical}, which is defined as: \begin{small}  $\mathbf{c}^k{=}\frac{1}{N^k}\sum_{i=1}^{N^k}\mathbf{z}_j$ \end{small}, where \begin{small}$N^k$\end{small} denotes the number of samples in class \begin{small}$k$\end{small}, $\mathbf{z}_j$ is the feature from $\mathcal{M}$ . 

After calculating all prototype representations \begin{small}$\{\mathbf{c}^k|\forall k\in[1,K]\}$\end{small}, which serve as the vertices in the prototype graph \begin{small}$\mathcal{P}_i$\end{small}, we further need to define the edges and the corresponding edge weights. 
The edge weight $A_{\mathcal{P}}(\mathbf{c}^i,\mathbf{c}^j)$ between two prototypes \begin{small}$\mathbf{c}^i$\end{small} and \begin{small}$\mathbf{c}^j$\end{small} is gauged by the similarity between them. The edge weight is calculated as follows:
\begin{equation}
\label{eq:proto_weight}
    A_{\mathcal{P}}(\mathbf{c}^i,\mathbf{c}^j)=\sigma(\mathbf{W}_{p}(|\mathbf{c}^i-\mathbf{c}^j|/\gamma_p)+\mathbf{b}_{p}),
\end{equation}
where \begin{small}$\mathbf{W}_{p}$\end{small} and \begin{small}$\mathbf{b}_{p}$\end{small} are the learnable parameters, \begin{small}$\gamma_p$\end{small} is a scalar. 
Thus, we denote  the prototype graph as \begin{small}$\mathcal{P}{=}(\mathbf{C}_{\mathcal{P}}, \mathbf{A}_{\mathcal{P}})$\end{small}, where \begin{small}$\mathbf{C}_{\mathcal{P}_i}{=}\{\mathbf{c}^i|\forall i\in[1,K]\}\in\mathbb{R}^{K\times d}$\end{small} represent a set of vertices, with each one corresponding to the prototype from a class, while \begin{small}$\mathbf{A}_{\mathcal{P}}{=}\{|A_{\mathcal{P}}(\mathbf{c}^i,\mathbf{c}^j)|\forall i,j\in[1,K]\}\in\mathbb{R}^{K\times K}$\end{small} gives the adjacency matrix, which indicates the proximity between prototypes.
We then use the prototype graph to guide the learning of the meta super-class graph. We construct a super graph $\mathcal{S}$ by connecting  prototype graph $\mathcal{P}$ to  super-class graph $\mathcal{C}$. In the super graph $\mathcal{S}$, the vertices are  $\mathbf{M}{=}(\mathbf{C}^l_{\mathcal{P}^l}; \mathbf{H}^l_{\mathcal{C}^l})$, where $\mathcal{P}^l$ denotes the refined prototype graph vertex after the $l$-th level super-class graph. 
Then, we calculate the link weight $A^l_{\mathcal{S}}(c^i, \{\mathbf{h}^j\})$ of the super graph as:
\begin{equation}
\label{eq:intra_sim}
    A^l_{\mathcal{S}}(\mathbf{c}^i, \mathbf{h}^{j^l})=\frac{\exp(-\Vert(\mathbf{c}^i-\mathbf{h}^{j^l})/\gamma_s^l\Vert_2^2/2)}{\sum_{{j^l}^{'}=1}^J \exp(-\Vert(\mathbf{c}^i-\mathbf{h}^{{j^l}^{'}})/\gamma^l_s\Vert_2^2/2)},
\end{equation}
where $\gamma_s^l$ is a scaling factor.  Note that, here we use  softmax to ensure that the total weight of edges between the prototype graph $\mathcal{P}$ and the super-class graph $\mathcal{C}$ is equal to 1, giving the prototype graph a unique influence on the expression of each super-class. Thus, the adjacent matrix and feature matrix of the super graph \begin{small}$\mathcal{S}^l{=}(\mathbf{A}^l,\mathbf{M}^l)$\end{small} is defined as \begin{small}$\mathbf{A}^l{=}(\mathbf{A}_{\mathcal{P}},\mathbf{A}^l_{\mathcal{S}};{\mathbf{A}^l_{\mathcal{S}}}^\mathrm{T},\mathbf{A}^l_{\mathcal{C}})$\end{small} and \begin{small}$\mathbf{M}^l{=}(\mathbf{C}^l_{\mathcal{P}^l};\mathbf{H}^l_{\mathcal{C}^l})$\end{small}. 
Once we constructed the super graph $\mathcal{S}$, we use message-passing again to propagate the most relevant knowledge from the prototype graph $\mathcal{P}$ to the super-class graph $\mathcal{C}$. Similar to eq.~(\ref{eq:mpc}):
\begin{equation}
    \mathbf{M}^{(m+1)}_{\mathcal{}}=\texttt{MP}(\mathbf{A}^l,\mathbf{M}^{(m)};\mathbf{W}^{(m)}).
    \label{eq:hmpc}
\end{equation}
We leverage the graph $\mathcal{S}$ to refine the super-class graph. Finally, we feed the original feature $\mathbf{z}$ into the super-class graph to get the information-propagated feature representation $\tilde{\mathbf{z}}^L$, which refines the original feature by its corresponding discovered super-classes. We provide the complete SuperDisco and Meta-SuperDisco algorithm specifications in the supplemental material.

\section{Experiments}
\label{sec:exp}

\textbf{Datasets.} 
%
We apply our method to four commonly used long-tail recognition benchmarks. Sample images and the number of categories for all datasets are provided in the supplement material.
\textit{\textbf{CIFAR-100-LT}} reduces the number of training samples per class according to an exponential function $n {=} n_i \mu^i$, where $i$ is the class index, $n_i$ is the original number of training samples, and $\mu \in (0,1)$. The imbalance factor of a dataset is deﬁned as the number of training samples in the most populated class divided by the minority class. We consider imbalance factors $\{10, 50, 100\}$. 
 \textit{\textbf{ImageNet-LT}}~\cite{liu2019large}  is a subset of ImageNet~\cite{deng2009imagenet} consisting of 115.8K images from 1000 categories, with maximally 1,280 images per class and minimally 5 images per class, and a balanced test set.  
 \textit{\textbf{Places-LT}}~\cite{liu2019large} has an imbalanced training set with 62,500 images for 365 classes from Places~\cite{zhou2017places}. It contains images from 365 classes and the number of images per class ranges from 4980 to 5. The test sets are balanced and contain 100 images per class.
%
 \textit{\textbf{iNaturalist}}~\cite{van2018inaturalist} is a real-world long-tailed dataset with  675,170 training images for  5,089 classes, where the top 1\% most populated classes contain more than 16\% of the training images.  Additionally, there is also a severe imbalance among the super-classes of iNaturalist. The 13 ground truth super-classes images range from 158,407 to 308.

\textbf{Implementation details.}
We follow~\cite{kang2019decoupling} by first training a feature extractor with instance-balanced sampling, and then training our graph model and classifier based on the trained features. For  CIFAR-100-LT, we follow~\cite{shu2019meta} and use a ResNet-32  backbone. For   ImageNet-LT, we use ResNeXt-50~\cite{he2016deep} as our backbone, following~\cite{kang2019decoupling}.  For Places-LT, we report results with ResNet-152 following~\cite{liu2019large}. For iNaturalist, we use a ResNet-50 backbone. We train each dataset for 200 epochs with batch size 512. We  use random left-right flipping and cropping as our training augmentation. 
For all experiments, we use an SGD optimizer with a momentum of 0.9 and a batch size of 512. We randomly selected 10 images per class from the training set for all datasets as $\mathcal{M}$. Code  available at:~\url{https://github.com/YDU-uva/SuperDisco}.

\begin{table}[t]
	\centering
	\scalebox{0.95}{\begin{tabular}{l c c c cc}
			\toprule
			  &  \multicolumn{5}{c}{Imbalance ratio}\\
			  	\cmidrule(lr){2-6}
			&  10 & 20 & 50 & 100 & 200 \\
			\midrule
		Baseline & 60.3 & 57.3  & 47.5 &44.9& 39.3\\
					\midrule
			SuperDisco & 65.9 & 60.7 & 57.2 & 50.9 & 45.2 \\
			Meta-SuperDisco & 68.5 & 63.1 & 58.3 & 53.8 & 47.5  \\
			\bottomrule
		\end{tabular}}
\caption{\textbf{Benefit of  SuperDisco and  Meta-SuperDisco.}  SuperDisco achieves better performance compared to a baseline fine-tuning on all imbalance factors, while  Meta-SuperDisco is even better for
long-tailed recognition.}
\label{ab_1}
\end{table}

\textbf{Benefit of SuperDisco and Meta-SuperDisco.}
To show the benefit of SuperDisco, we compare it with a fine-tuning baseline, which retrains the classifier only.  Table \ref{ab_1} shows SuperDisco improves over fine-tuning  on CIFAR-100-LT, and the results for the other long-tailed datasets are provided in the supplemental materials Table~1. In the most challenging setting with the largest imbalance factor of $200$, our SuperDisco delivers $45.2\%$, surpassing the baseline by $5.9\%$.
We attribute improvement to our model's ability to refine  original features,
allowing the discovered super-class graph to guide the tail features away from the dominant role of  head features, thus leading to improvements over the original features.  We also investigate  the benefit of meta-learning with  Meta-SuperDisco. The  Meta-SuperDisco consistently surpasses the  SuperDisco for all imbalance factors. The consistent improvements confirm that  Meta-SuperDisco learns even more robust super-class graphs, leading to a discriminative representation of the tail data.

\begin{table}[t]
\centering
\scalebox{0.95}{\begin{tabular}{c c c c c c}
			\toprule
			   &  \multicolumn{5}{c}{Imbalance ratio}\\
			   \cmidrule(lr){2-6}
			 &  10 & 20 & 50 & 100 & 200 \\
            \midrule
		Baseline & 60.3 & 57.3  & 47.5 &44.9& 39.3\\
		\midrule
		(20) & 61.2 & 60.1 & 49.9 & 47.3& 41.9\\
		(2, 4, 8) &65.3 & 62.7 & 53.1 & 49.8&  43.2\\
		(4, 8, 16) & \textbf{69.1} & \textbf{64.2} & 55.2 & 52.3& 45.9\\
		(4, 8, 16, 32) &  68.5 & 63.1 & {58.3} & \textbf{53.8} & \textbf{47.5}\\
		(4, 8, 16, 32, 64) &  66.9 & 62.7 & \textbf{58.9} & {52.9} & {46.3}\\
		\midrule
		Oracle super-classes &  66.9  &63.2 & 54.7 & 51.4 & 43.2\\
			\bottomrule
		\end{tabular}
}
\caption{\textbf{Effect of number of super-class levels on CIFAR-100-LT}.  Compared to a baseline\cite{kang2019decoupling} and an oracle setting, Meta-SuperDisco provides higher performance gains with more complex hierarchies.}
\label{ab_3}
\end{table}

\begin{table}[t]
\centering
\scalebox{0.8}
{\begin{tabular}{c c c c c}
			\toprule
			 &  Many & Medium & Few & All \\
            \midrule
		Baseline & 65.0 & 66.3 & 65.5 & 65.9 \\
		\midrule
		(13) & 71.8  & 70.2 & {66.1} & 70.8\\
		(2, 4, 8) & 70.5 & 69.3 & {65.9} & 69.4\\
		(4, 8, 16)  & 72.2 & 70.9 & {66.4} & 70.3\\
		(4, 8, 16, 32) & \textbf{73.6} & 70.2 & {67.3} & 70.9\\
		(4, 8, 16, 32, 64) & {73.4} & \textbf{72.9} & \textbf{68.3} & \textbf{72.3} \\
	    (4, 8, 16, 32, 64, 128) & 72.1 & 71.3 & {66.2} & 70.9 \\
		\midrule
		Oracle super-classes & 70.7  & 70.5 & {65.9} & 70.2\\
			\bottomrule
		\end{tabular}
}
\caption{\textbf{Effect of number of super-class levels on iNaturalist}.  Meta-SuperDisco achieves consistent performance gains with more complex hierarchies.}
\label{ab_3_inau}
\end{table}

\textbf{Effect of the number of super-class levels.} 
A significant challenge with any structure-aware learning algorithm is determining the appropriate complexity for the knowledge structure. So, we further analyze the effect of the super-class hierarchies, including the level (number of depths L) or the number of super-classes in each level. The results are shown in Table~\ref{ab_3} and Table~\ref{ab_3_inau}. The super-class number from the bottom layer to the top layer is saved in a tuple.
For example, (2, 4, 8) represents three depth, with two super-classes in the top layer. The baseline is Decouple-LWS~\cite{kang2019decoupling}, which only inputs  the original feature to learn a new classifier. 
The oracle super-classes  are first trained on two long-tailed datasets using the ground truth super-class labels for super-class classifications.
Once the training is completed, each oracle super-class is obtained by averaging the samples of each super-class.
We constructed a one-layer super-class graph using these super-classes,
where the vertices of the graph are for each super-class, and the edges of the graph are computed according to Eq.~(\ref{eq:hc_e}). 
Then, we use the message passing by Eq.~(\ref{eq:mpc}) to refine the original features and input them into the classifier to get the final predictions.
From Table~\ref{ab_3},  we observe that using oracle super-classes achieves better performance compared to the learned super-class (20) since it uses the ground truth super-classes as supervision. We also conclude that too few levels may not be enough to learn the precise  super-classes (\eg, tuple (20) or (2, 4, 8)).  In this dataset, increasing levels (\eg, tuple (4, 8, 16, 32)) achieves better performance on the smaller imbalance factor (\eg, 10), and similar performance compared with (4, 8, 16). 
For the real-world long-tailed dataset iNaturalist~\cite{VanHorn2017} in  Table~\ref{ab_3_inau}, we find no significant improvement for the few-shot classes in the performance of the oracle super-class compared to the baseline, and the same is true for the performance of the discovered super-class structure (13). 
This is because the super-class of iNaturalist also have serious long-tailed problems, resulting in the refined features of tail classes remaining indistinguishable from the refined features of head classes.
However, with a more complex graph structure (4, 8, 16, 32, 64), the few-shot performance improves by a good margin compared with the baseline, and even the oracle super-classes. We attribute this to our model's ability to explore relatively balanced super-class spaces, thus making the  refined tail category features  discriminative. By comparing Table~\ref{ab_3} and Table~\ref{ab_3_inau}, we conclude that deeper as well as wider graphs are needed to discover the super-classes in the case of severe class imbalance.

\begin{figure} [t]
  \begin{minipage}{0.5\columnwidth}
    \subfloat[SuperDisco]{
      \includegraphics[width=.95\columnwidth]{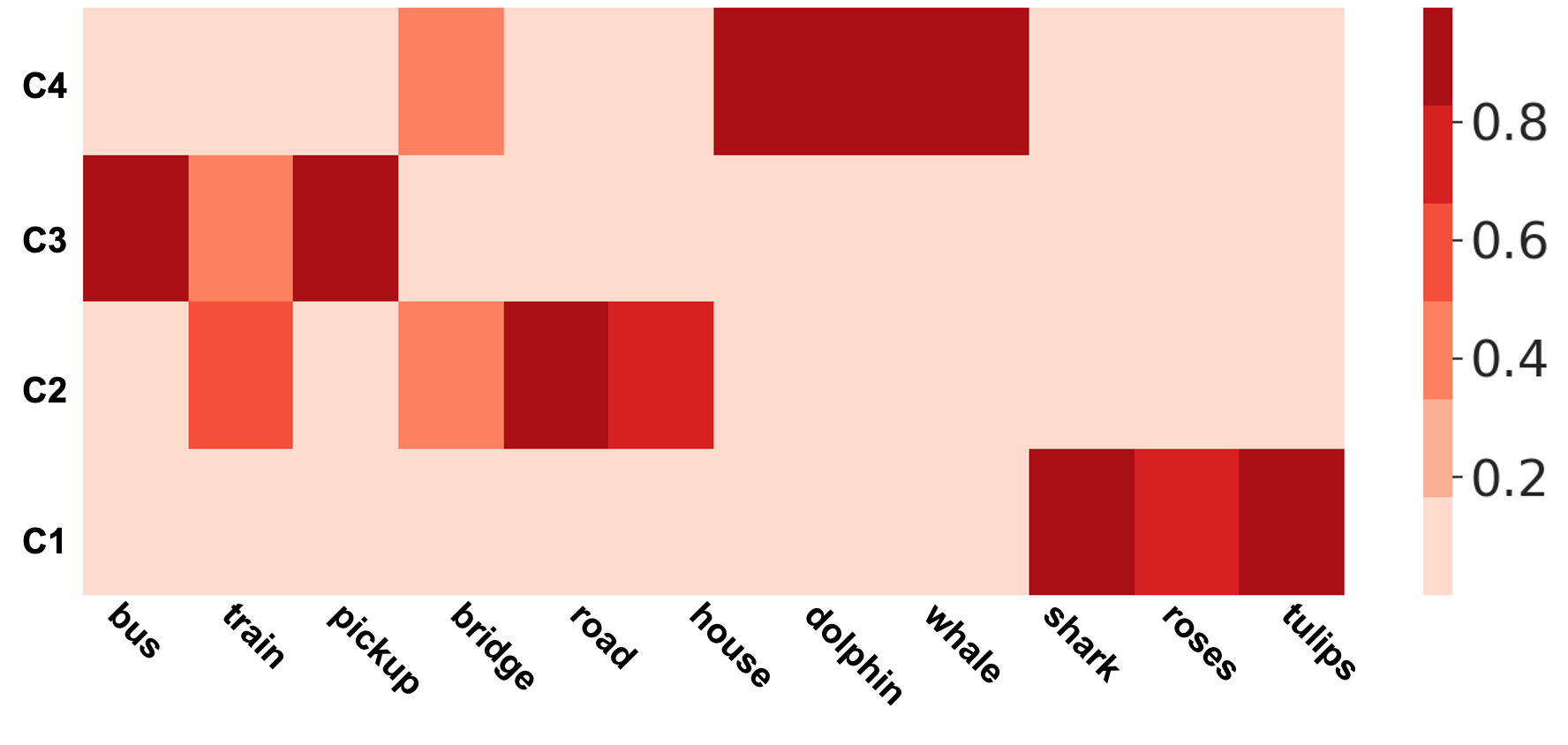}}
         \subfloat[Meta-SuperDisco]{
      \includegraphics[width=.95\columnwidth]{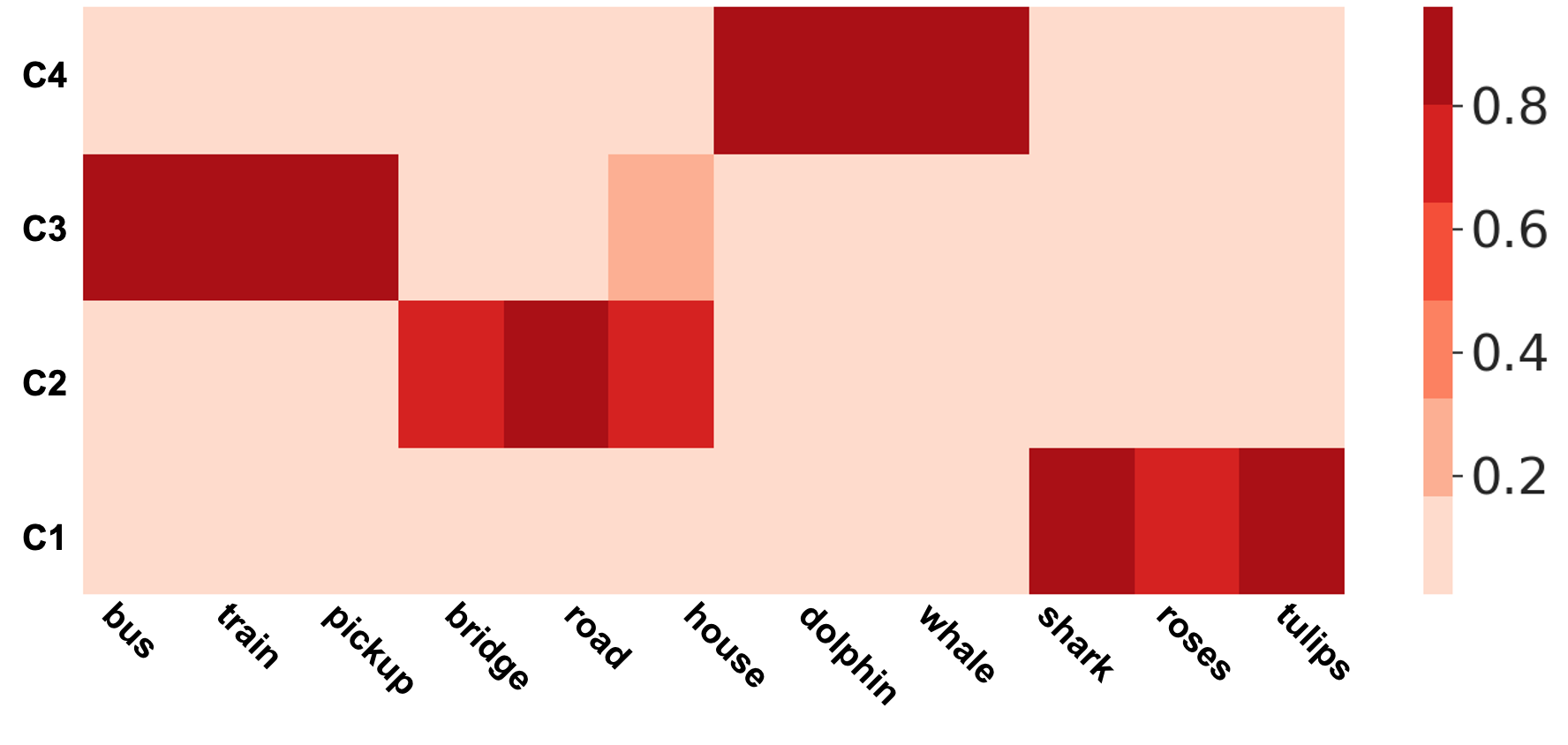}}
\end{minipage}
\caption{\textbf{Similarity between discovered super-classes and classes.} SuperDisco discovers super-classes hidden in each class, while Meta-SuperDisco discovers  more accurate super-classes. }\label{fig:cg}
\vspace{-2mm}
\end{figure}

\textbf{Visualization of  SuperDisco.}
To understand the meaning of the discovered super-classes more clearly, we present a visualization in Figure.~\ref{fig:cg}. We selected 12 different categories from the CIFAR-100 test dataset. We calculate the similarity of each of these 12 categories to the different vertices in the graph we explore. Here we show the similarity with the second layer of graph vertices (C1, C2, C3, C4).  We can see different categories mainly activate different vertices, e.g., bus $\rightarrow$  C3 and road $\rightarrow$ C2. As shown in this heatmap, we find that C1 reflects the super-class of \textit{flowers},  C2 reflects the super-class of \textit{buildings}, C3 reflects the super-class of \textit{vehicles}, C4 reflects the super-class of \textit{fish}.  Another observation is that the second-largest activated super-class is also meaningful, promoting knowledge transfer between super-classes.
For example,  \textit{road} and \textit{bridge} are related to the C3 super-class, since some vehicles may be on the road and bridge. 
This visualization reflects that we can use graph models to discover the super-classes and the relationships between each super-class.
We also visualize the discovered meta-learning super-classes in Figure~\ref{fig:cg} (b). The discovered super-classes are even more accurate, e.g., \textit{roses} have high similarity to C2, which mainly reflects the \textit{buildings} super-class, while it has high similarity to C1, which is the \textit{flowers} super-class.  This once again validates the benefit of Meta-SuperDisco. 
Furthermore, in the (c) learned the hierarchical concept of each class, we can see that bus and bridge have the same concept C2 in the last concept level, which may be due to the possible presence of cars on the bridge.

\begin{figure}[t]
\centering
\includegraphics[width=.6\linewidth]{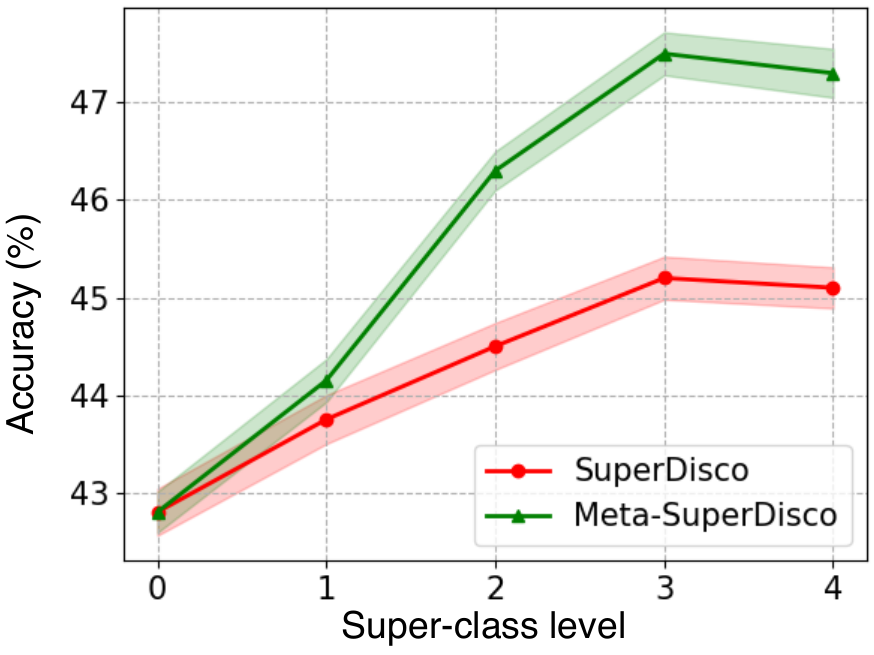}
\caption{\textbf{Effect of refined features.} Accuracy increases along with the increased super-class levels, revealing that more accurate and richer super-classes facilitate better long-tailed recognition.}
\label{fig:level}
\end{figure}

\begin{figure*} [t]
\centering
\includegraphics[width=0.9\linewidth]{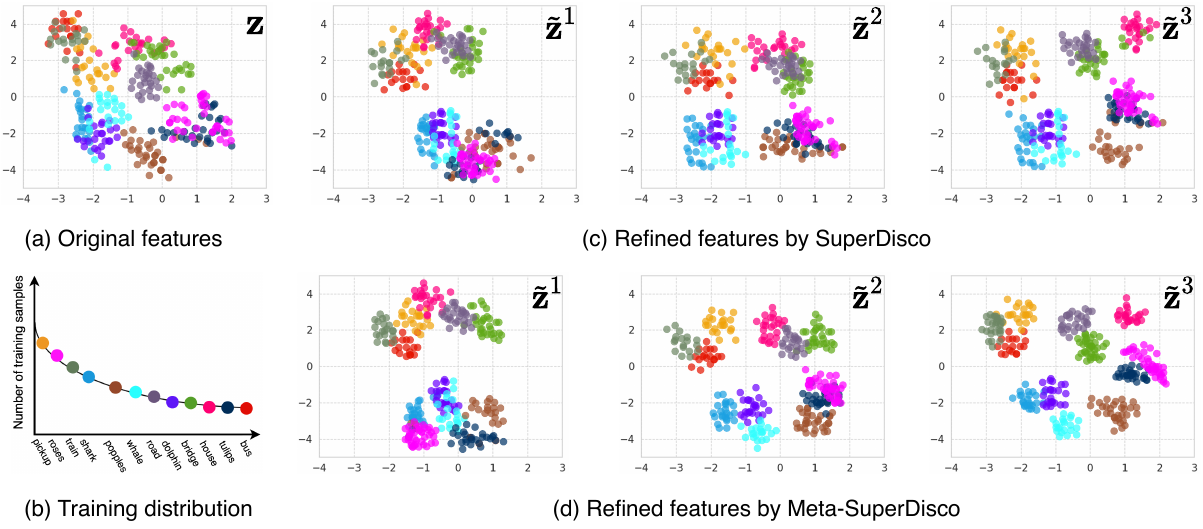}
 \caption{\textbf{Visualization of refined features} on CIFAR-100-LT, with the original features (a) and their corresponding training distribution (b). Colours indicate categories. SuperDisco (c) guides the original features on being clustered into the corresponding super-class space at different levels, while Meta-SuperDisco (d) obtains even more discriminative intra-class features.}
 \label{fig:tsne}
\end{figure*}

\textbf{Visualization of refined features.}
To understand the empirical benefit of SuperDisco,
we visualize in Figure~\ref{fig:tsne} the original features and refined features 
with super-class graphs of the different levels using t-SNE~\cite{van2008visualizing}. We choose the vertices numbers as (2, 4, 6), meaning the super-class graph has three different levels, each with a different number of vertices.
The original features of the category with a small sample size will overlap with the (original) features of the category with a large sample size. Super-class graphs  discovered by our model  project the original features into a high-level super-class space, where the imbalance is relatively small. Hence, its corresponding subcategory can be predicted more accurately. 
It is worth noting that when comparing the two different super-class graphs on top and below, the features obtained by Meta-SuperDisco are even more distinctive and 
distant from each other.
To better measure the goodness of the refined features obtained at different levels, we show in Figure~\ref{fig:level} the prediction accuracy using different refined 
features. We find that the accuracy increases along with the
increased super-class levels, which shows that using more accurate and richer super-classes facilitates better performance. This again demonstrates that Meta-SuperDisco is most suitable for long-tailed visual recognition. 

\begin{table}[t]
\centering
\scalebox{0.9}
{\begin{tabular}{l c c c c}
			\toprule
			 &  Many & Medium & Few & All \\
            \midrule
            Baseline  & 58.4  & 49.3 & 34.8 & 52.7\\
		Multi-layer perceptron & 63.5  & 51.8 & 35.9 & 55.0\\
  \rowcolor{Gray}
		Graph convolution network & 66.1  & 53.3 & 37.1 & 57.1\\
			\bottomrule
		\end{tabular}
}
\caption{\textbf{Analysis of super-class mechanism} on ImageNet-LT. The super-class mechanism contributes most, the graph convolution network improves results further.}
\label{ab_GCN}
\end{table}

\begin{table*}[t]
 \centering
\scalebox{0.8}
{\begin{tabular}{l l c c c c  c c c c  c c c c }
\toprule
&  &  \multicolumn{4}{c}{\textit{\textbf{ImageNet-LT}}} &   \multicolumn{4}{c}{\textit{\textbf{Places-LT}}}  &   \multicolumn{4}{c}{\textit{\textbf{iNaturalist}}}\\
\cmidrule(lr){3-6} \cmidrule(lr){7-10} \cmidrule(lr){11-14} 
& Venue & Many & Medium & Few & All  & Many & Medium & Few & All  & Many & Medium & Few & All \\ 
\midrule 
Kang~\etal~\cite{kang2019decoupling} & ICLR 19 &60.2 & 47.2 & 30.3 & 49.9 & 40.6 &	39.1 &	28.6 & 37.6 & 65.0 & 66.3 & 65.5 & 65.9 \\
Kang~\etal~\cite{kang2021exploring} & ICLR 21 & 61.8 & 49.4 & 30.9 & 51.5  &-&-&-&-  & -  & - & - & 68.6\\
He~\etal~\cite{he2021distilling}  & ICCV 21 & 64.1 & 50.4 & 31.5  & 53.1 &-&-&-&- & 70.6 & 70.1 & 67.6 & 69.1 \\ 
Li~\etal~\cite{li2021self} & CVPR 21 &\textbf{66.8} & 51.1 & 35.4  & 56.0 &-&-&-&-  & - & - & - & 69.3\\
Samuel~\etal~\cite{samuel2021distributional} & ICCV 21 & 64.0 & 49.8 & 33.1  & 53.5 &-&-&-&- & - & - & - & 69.7 \\ 
Alshammari~\etal~\cite{alshammari2022long} & CVPR 22 &62.5 & 50.4 & 41.5  & 53.9 &-&-&-&-  & 71.2 & 70.4 & 69.7 & 70.2\\
Zhang~\etal~\cite{zhang2021distribution} & CVPR 21 &61.3 & 52.2 & 31.4  & 52.9 &40.4&\textbf{\underline{\emph{42.4}}}&30.1 &39.3 & 69.0 & 71.1 & \textbf{\underline{\emph{70.2}}} & 70.6 \\ 
Parisot~\etal~\cite{parisot2022long}& CVPR 22 & 63.2 & 52.1 & \textbf{\underline{\emph{36.9}}}  & 54.1 &39.7&41.0&\textbf{\underline{\emph{34.9}}}& \textbf{\underline{\emph{39.2}}} & - & - & - & - \\
Park~\etal~\cite{park2022majority} & CVPR 22 &\textbf{\underline{\emph{{66.4}}}} & \textbf{53.9} & 35.6  & \textbf{\underline{\emph{56.2}}} &-&-&-&-  & \textbf{73.1} & \textbf{\underline{\emph{72.6}}} & 68.7 & \textbf{\underline{\emph{72.8}}}\\
	\rowcolor{Gray}
\textit{\textbf{This paper}} &  &66.1 & \textbf{\underline{\emph{53.3}}} & \textbf{37.1} & \textbf{57.1}  & \textbf{45.3} & \textbf{42.8} & \textbf{35.3} & \textbf{40.3}   & \textbf{\underline{\emph{72.3}}} & \textbf{72.9} & \textbf{71.3} & \textbf{73.6}
\\
\bottomrule
\end{tabular}}
 \caption{\textbf{Comparison with the state-of-the-art on ImageNet-LT, Places-LT and iNaturalist.} Best and second best results are highlighted in \textbf{bold} and \textbf{\underline{\emph{italic bold}}}.  Our Meta-SuperDisco achieves either better or comparable performance than state-of-the-art methods under the tail and all data for long-tailed visual recognition.}
  \label{tab:imagenet-pl}
  \vspace{-5mm}
\end{table*}
\begin{table}[t]
\centering
\scalebox{0.8}
{\begin{tabular}{l l c c c}\toprule 
        &  & \multicolumn{3}{c}{Imbalance ratio}\\
                \cmidrule(lr){3-5}
        & Venue & 10 &50 &100\\
        \midrule
        Park~\etal~\cite{park2021influence} & ICCV 21 & 59.5	 	& 47.4 & 42.0\\
        Li~\etal~\cite{li2021self} & CVPR 21  &62.3      & 50.5	 	& 46.0 \\
        Zhong~\etal~\cite{zhong2021improving}& CVPR 21  & 62.5     & 51.5 	& 46.8  \\ 
        Samuel~\etal~\cite{samuel2021distributional}& ICCV 21  &  63.4	 	& 57.6	& 47.3	 \\
        Wang~\etal~\cite{wang2020long} & ICLR 21 & 61.8       &51.7	 	& 48.0 \\
        Zhu~\etal~\cite{zhu2022balanced} & CVPR 22  &64.9       & 56.6	 	& 51.9 \\
        Cui~\etal~\cite{cui2021parametric}  & ICCV 21 & 64.2	 	& 56.0	& 52.0	 \\
        Alshammari~\etal~\cite{alshammari2022long} & CVPR 22  &\textbf{\underline{\emph{68.8}}}     &\textbf{\underline{\emph{57.7}}}	 	& \textbf{\underline{\emph{53.3}}} \\
         \midrule
        	\rowcolor{Gray}
        \textit{\textbf{This paper}}  &  &\textbf{69.3}      & \textbf{58.3}	 	&  \textbf{53.8} \\
        \bottomrule
	\end{tabular}}
\caption{\textbf{Comparison with the state-of-the-art on
CIFAR-100-LT.} Our model achieves best performance. }
\label{tab:cifar}
\vspace{-2mm}
\end{table} 
\textbf{Analysis of super-class mechanism.}
To demonstrate that the improved performance of our SuperDisco cannot solely be attributed to the graph convolutional network module, we conducted an experiment where we replaced it with a multi-layer perceptron to obtain the representation per sample.  In Table~\ref{ab_GCN}, the performance gains of our method are primarily due to the super-classes rather than the graph convolution network. 
The results suggest that incorporating the super-classes mechanism plays a crucial role in improving the performance of long-tailed problems. Furthermore, the results improve further when we replace the multi-layer perceptron with our graph convolution network module.

\textbf{Comparison with the state-of-the-art.}
We evaluate our method on the four long-tailed datasets under different imbalance factors in Table~\ref{tab:imagenet-pl} and \ref{tab:cifar}. Our model achieves state-of-the-art performance on the tail data of all datasets.  
For ImageNet-LT, our model achieves state-of-the-art performance on both few-shot and all data. In the most challenging Places-LT, our model delivers $40.3\%$ on all classes, surpassing the second-best Parisot~\etal~\cite{parisot2022long} by $1.1\%$. On the real-world long-tailed dataset iNaturalist, our model achieves the three best performances under four different shots. 
On the long-tailed synthetic dataset CIFAR-100-LT, our model achieves the best performance under each imbalance factor.
The consistent improvements on all benchmarks under various conﬁgurations conﬁrm that our Meta-SuperDisco is effective for long-tailed visual recognition.

\begin{figure}[t]
\centering
\includegraphics[width=0.7\linewidth]{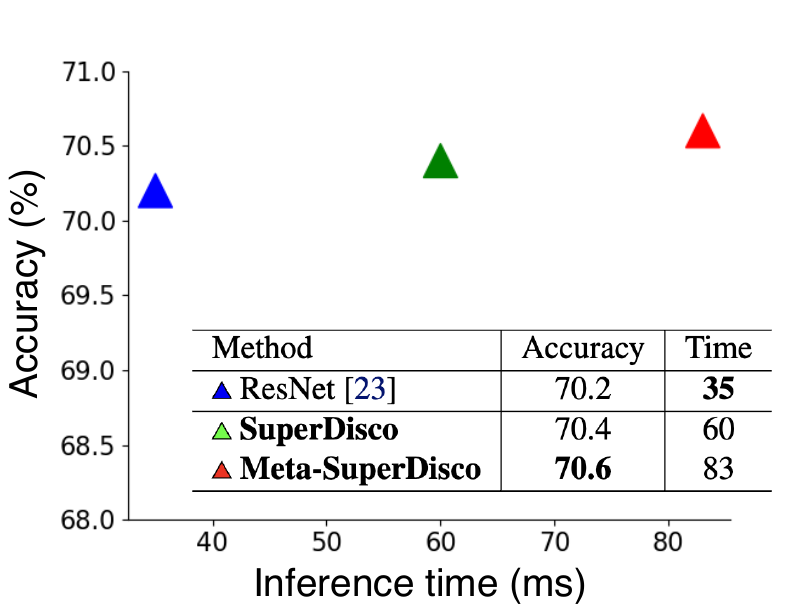}
    \caption{\textbf{Limitation.} Accuracy (\%) vs. speed (ms) comparison with different methods on balanced CIFAR-100. SuperDisco  has little impact on the performance of balanced datasets at the expense of increased inference time.}
\label{fig:speed}
\vspace{-6mm}
\end{figure}

\textbf{Limitations.} 
We show that SuperDisco and Meta-SuperDisco achieve good performance on tail data while 
being less successful on the head data. Based on this result, we also perform an experiment on \textit{balanced} CIFAR-100 in Figure~\ref{fig:speed}. 
With SuperDisco and Meta-SuperDisco, there is only a slight change in performance at the expense of an increased inference time. This reveals that our SuperDisco does not change the original features much through message passing on a balanced dataset. This may be because the obtained super-classes are still the original class itself. In addition, as the computation of graphs involves many matrix operations, our model also requires a relatively long computational speed.  Due to introducing a prototype graph and more data, Meta-SuperDisco takes longer to compute.  In addition, the training time of SuperDisco and its meta variants is also 1.5  times higher than the baseline.
Future work could investigate how to use the discovered super-class graph in balanced datasets and how to reduce the computation time.

\section{Conclusions}

This paper proposes learning to discover a super-class graph for long-tailed visual recognition.
The proposed super-class graph could rectify and refine the original features by message passing, which results in attending to the most relevant entities based on their semantic similarity between concepts for more accurate predictions. 
To obtain a more informative super-class graph and more balanced image representations, we further propose to meta-learn the super-class graph based on the prototype graph from a small amount of imbalanced data.
We conduct thorough ablation studies to demonstrate the effectiveness of the proposed SuperDisco and Meta-SuperDisco algorithms. 
The state-of-the-art performance on the long-tailed version of four datasets further substantiates the beneﬁt of our proposal.

\paragraph{Acknowledgment}
This work is financially supported by the Inception Institute of Artificial Intelligence, the University of Amsterdam and the allowance 
Top consortia for Knowledge and Innovation (TKIs) from the Netherlands Ministry of Economic Affairs and Climate Policy.

\medskip


{\small
\bibliographystyle{ieee_fullname}
\bibliography{arxiv}
}

\clearpage
\appendix

\section{Effect of number of super-class levels on more datasets.} 
\begin{figure}[t]
\centering
\includegraphics[width=0.7\linewidth]{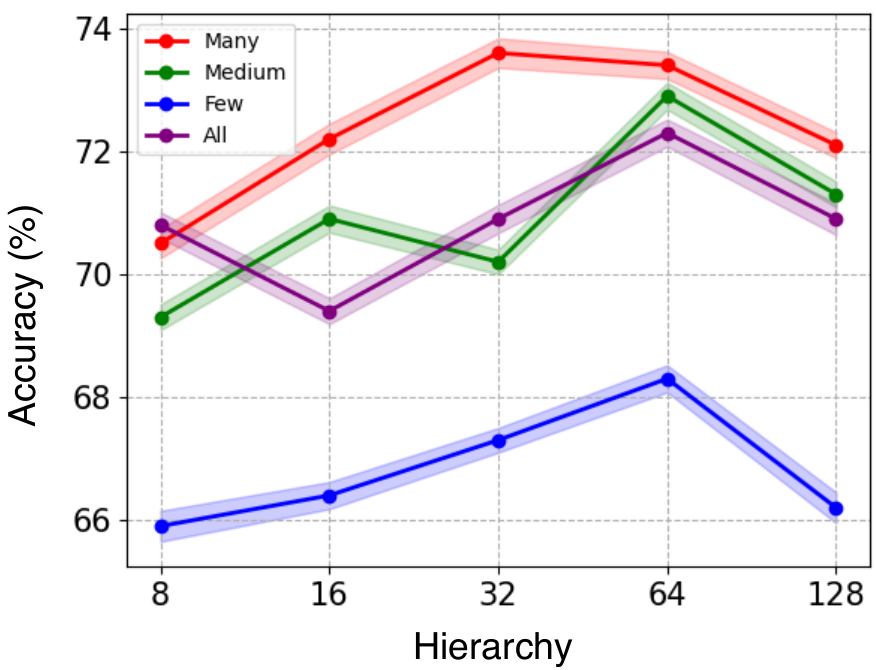}
\caption{\textbf{Effect of number of super-class levels} on {iNaturalis}-LT. }
\label{fig:ina_ana}
\end{figure}

We experimented with \textit{\textbf{ImageNet-LT}}~\cite{liu2019large}  and \textit{\textbf{Places-LT}}~\cite{liu2019large} with different number of super-class levels. The results are reported in Table~\ref{ab_3_imagenet} and Table~\ref{ab_3_place}, respectively. 
On the \textit{\textbf{ImageNet-LT}}~\cite{liu2019large}, we find that the performance of the super-class graphs with the different number of super-class levels is higher than the baseline. However, with more hierarchies (\ie{the last row}), the performance on the few-shot classes is the highest, while (4, 8, 16, 32, 64) achieves the best performance on all classes.  
On the \textit{\textbf{Places-LT}}~\cite{liu2019large}, with more complex hierarchies \ie{(4, 8, 16, 32, 64, 128, 258)} achieves the best performance on all classes and few-shot classes.  
We also conduct experiment on the \textit{\textbf{iNaturalis}}~\cite{van2018inaturalist} to analysis the effect of number of super-class levels in the Figure~\ref{fig:ina_ana}. We can find that with more hierarchies, the performance will consistently increase. 64 achieves the peak performance on the all classes and any-shot classes.
For this experiment, we attribute this to our model's ability to explore relatively balanced super-class spaces, thus making the refined tail category features discriminative. We conclude that deeper and broader graphs are needed to discover the super-classes in the case of severe class imbalance.

\begin{table}[h]
\centering
\scalebox{0.9}
{\begin{tabular}{c c c c c}
			\toprule
			 &  Many & Medium & Few & All \\
            \midrule
		Baseline & 57.1 & 45.2 & 29.3 & 47.7 \\
		\midrule
		(2, 4, 8) & 58.6 & 47.1 & 31.1 & 49.8\\
		(4, 8, 16)  & 59.8 & 48.3 & 33.2 & 50.1\\
		(4, 8, 16, 32)  & 61.3 & 49.7 & 35.1 & 52.9\\
	    (8, 16, 32, 64)  & \textbf{66.5} & 49.8 & 36.1 & 55.1\\
		(4, 8, 16, 32, 64) & 66.4 & \textbf{53.3} & 37.1 & \textbf{57.1} \\
	    (4, 8, 16, 32, 64, 128) & 66.1 & 52.3 & \textbf{37.9} & 56.5 \\
			\bottomrule
		\end{tabular}
}
\caption{\textbf{Effect of number of super-class levels on \textit{\textbf{ImageNet-LT}}}.  Meta-SuperDisco achieves consistent performance gains with more complex hierarchies.}
\label{ab_3_imagenet}
\end{table}

\begin{table}[ht]
\centering
\scalebox{0.8}
{\begin{tabular}{c c c c c}
			\toprule
			 &  Many & Medium & Few & All \\
            \midrule
		Baseline & 40.6 & 39.1 & 28.6 & 37.6 \\
		\midrule
		(2, 4, 8) & 43.1 & 39.1 & {29.9} & 37.5\\
		(4, 8, 16) & 44.2 & 39.9 & {30.3} & 38.1\\
		(4, 8, 16, 32) & \textbf{45.9} & 40.4 & {31.1} & 38.9\\
		(4, 8, 16, 32, 64)  & 44.9 & 41.3 & {32.3} & 39.2\\
	    (4, 8, 16, 32, 64, 128)  & 44.3 & \textbf{43.1} & {34.5} & 39.9\\
	    (4, 8, 16, 32, 64, 128, 256) & 45.3 & 42.8 & \textbf{35.3} & \textbf{40.3}\\
         (4, 8, 16, 32, 64, 128, 256, 512) & 44.1 & 42.3 & {34.0} & \textbf{39.1}\\
			\bottomrule
		\end{tabular}
}
\caption{\textbf{Effect of number of super-class levels on \textit{\textbf{Places-LT}}}.  Meta-SuperDisco achieves consistent performance gains with more complex hierarchies.}
\label{ab_3_place}
\end{table}

\section{Benefit of  SuperDisco and  Meta-SuperDisco} 
We also give the ablation to show the benefit of SuperDisco and  Meta-SuperDisco on \textit{\textbf{ImageNet-LT}}/\textit{\textbf{Places-LT}}/\textit{\textbf{iNaturalist}} in Table~\ref{tab:ab_all}.   The Meta-SuperDisco consistently surpasses the SuperDisco for all shots. The consistent improvements confirm that Meta-SuperDisco learns even more robust super-class graphs, leading to a discriminative representation of the tail data.

\begin{table*}[ht]
 \centering
\scalebox{1.}
{\begin{tabular}{l l c c  c  c c c c  c c c c }
\toprule
  &  \multicolumn{4}{c}{\textit{\textbf{ImageNet-LT}}} &   \multicolumn{4}{c}{\textit{\textbf{Places-LT}}}  &   \multicolumn{4}{c}{\textit{\textbf{iNaturalist}}}\\
\cmidrule(lr){2-5} \cmidrule(lr){6-9} \cmidrule(lr){10-13} 
 & Many & Medium & Few & All  & Many & Medium & Few & All  & Many & Medium & Few & All \\ 
\midrule 
 Baseline & 58.4 & 49.3 & 34.8  & 52.7 &42.1&39.2&30.9& 36.3 & 68.3 & 69.2 & 67.1 & 68.5 \\
 \midrule 
SuperDisco & 65.1 & 52.1 & 35.9  & 55.9  &44.7&41.1&34.2&39.2  & {71.3} & 71.0 & 69.6 & 72.1\\
Meta-SuperDisco  &66.1 & 53.3 & {37.1} & {57.1}  & {45.3} & {42.8} & {35.3} & {40.3}   & 72.3 & {72.9} & {71.3} & {73.6}
\\
\bottomrule
\end{tabular}}
\caption{\textbf{Benefit of  SuperDisco and  Meta-SuperDisco.}  SuperDisco achieves better performance compared to a baseline fine-tuning on all shots, while  Meta-SuperDisco is even better for long-tailed recognition.}
  \label{tab:ab_all}
\end{table*}

\section{Computation cost}
We report the computation cost and accuracy gain ablation in Table~\ref{flops} for ImageNet-LT. Although our model requires more parameters and computational costs compared to the baseline, it brings a 7.2\% improvement in accuracy. Compared to the state-of-the-art method by Park \etal~\cite{park2022majority}, our model requires a considerably lower amount of additional parameters and computational cost while still delivering better results. 

\begin{table}[ht]
\caption{\textbf{Computation cost and accuracy gain} for SuperDisco on ImageNet-LT compared to the baseline and state-of-the-art. SuperDisco provides a good trade off.}
\centering
\label{flops}
{%
\resizebox{1\linewidth}{!}{
\begin{tabular}{lrrr}
\toprule
& \multicolumn{2}{c}{Added computational cost} & \\
\cmidrule(){2-3}
Models  & FLOPs (M) & Parameters (M) & Accuracy\\
\midrule
ResNet-32  & 0 & 0 & 45.3\\ 
\hline
Baseline  & 0.04 & 0.001 & 49.9\\
Park et al~\cite{park2022majority} & 0.41 & 0.35 & 56.2\\
\midrule
\rowcolor{Gray}
SuperDisco  & 0.15 & 0.03 & 56.4\\
\rowcolor{Gray}
Meta-SuperDisco & 0.28 & 0.08 & 57.1\\
\bottomrule
\end{tabular}
}}
\end{table}

\section{Evaluation protocol} 
We evaluate our model on the test sets for each dataset and report commonly used top-1 accuracy over all classes. For the CIFAR-100-LT dataset, we report the accuracy with different imbalance factors. For the   \textit{\textbf{ImageNet-LT}},  \textit{\textbf{Places-LT}}, and  \textit{\textbf{iNaturalist}}, we follow~\cite{liu2019large} and further report accuracy on three different splits of the set of classes: \textit{Many-shot} ($>$100 images), \textit{Medium-shot} (20-100 images) and \textit{Few-shot} ($<$20 images). We report the average top-1 classiﬁcation accuracy 
across all test images.

\section{Algorithm}
We give the detailed algorithms of SuperDisco and Meta-SuperDisco in Alg.~\ref{alg:superdisco} and Alg.~\ref{alg:meta_superdisco}, respectively.

\begin{algorithm*}[ht]
\caption{{SuperDisco}}
\label{alg:superdisco}
\begin{algorithmic}[1]
\REQUIRE Training data: $\{x_k, y_k\}$;  Number of super-class levels: $l$;  Number of vertices in the $l$-th super-class level: $C^l$;  Feature extractor: $f_{\theta}(\cdot)$;  Graph function: $g_{\phi}(\cdot)$; Classifier function: $h_\psi(\cdot)$; Learning rate: $\alpha$.
\STATE Randomly initialize all learnable parameters $\Phi = \{\theta, \phi, \psi\}$
\WHILE{not done}
\STATE Sample a batch of samples $\{x_i, y_i\}$
\STATE Compute the original feature: $\mathbf{z} = f_{\theta}(x)$
\STATE Construct the super-class graph $\mathcal{C}^l$ by computing the super-class vertex $\mathbf{H}_{\mathcal{C}}^l$ and weights $\mathbf{A}_{\mathcal{C}}^l$ based on the Eq.~(1)
\STATE Construct the graph $\mathcal{R}$ and compute the weight $A^l_{\mathcal{R}}$ based on the Eq.~(2)
\FOR{m in the number of layers of GNN}
\STATE Apply GNN on the graph $\mathcal{R}$ by message passing and obtain the representations $\mathbf{H}^{(m+1)}_{\mathcal{R}}$ based on the Eq.~(3)
\ENDFOR
\STATE Get the refined feature $\mathbf{z^l} =  \mathbf{H}^{(m+1)}_{\mathcal{R}}[0]$
\STATE  Compute the final prediction $ \tilde{y} = h(\mathbf{z}^l)$
\STATE Update $\Phi =\Phi - \alpha \nabla_{\Phi} \sum^I_{i=1}\mathcal{L}_{\mathrm{CE}}(\tilde{y}_i, y_i)$
\ENDWHILE
\end{algorithmic}
\end{algorithm*}

\begin{algorithm*}[ht]
\caption{{Meta-SuperDisco}}
\label{alg:meta_superdisco}
\begin{algorithmic}[1]
\REQUIRE Training data: $\{x_k, y_k\}$;  Balanced data: $\mathcal{M}$;  Number of super-class levels: $l$;  Number of vertices in the $l$-th super-class level: $C^l$;  Feature extractor: $f_{\theta}(\cdot)$;  Graph function: $g_{\phi}(\cdot)$; Classifier function: $h_\psi(\cdot)$; Learning rate: $\alpha$.
\STATE Randomly initialize all learnable parameters $\Phi = \{\theta, \phi, \psi\}$
\WHILE{not done}
\STATE Sample a batch of samples $\{x_i, y_i\}$
\STATE Compute the original feature: $\mathbf{z} = f_{\theta}(x)$
\STATE Construct the super-class graph $\mathcal{C}^l$ by computing the super-class vertex $\mathbf{H}_{\mathcal{C}}^l$ and weights $\mathbf{A}_{\mathcal{C}}^l$ based on the Eq.~(1)
\STATE Construct the prototype graph $\mathcal{P}$ by computing the prototype vertex $\mathbf{C}_{\mathcal{P}}$ and weights $\mathbf{A}_{\mathcal{P}}$  based on the Eq.~(4)
\STATE Construct the graph $\mathcal{R}$ and compute the weight $A^l_{\mathcal{R}}$ based on the Eq.~(2)
\STATE Construct the super graph $\mathcal{S}$ and compute the vertices  $\mathbf{C}^l_{\mathcal{P}}$ and weight $\mathbf{H}^l_{\mathcal{C}^l}$ based on the Eq.~(5)
\FOR{m in the number of layers of GNN}
\STATE Apply GNN on the graph $\mathcal{S}$ by message passing and obtain the representations $\mathbf{M}^{(m+1)}$ based on the Eq.~(6)
\ENDFOR
\STATE Get the refined feature $\mathbf{z^l} = \mathbf{M}^{(m+1)}[0]$
\STATE  Compute the final prediction $ \tilde{y} = h(\mathbf{z}^l)$
\STATE Update $\Phi =\Phi - \alpha \nabla_{\Phi} \sum^I_{i=1}\mathcal{L}_{\mathrm{CE}}(\tilde{y}_i, y_i)$
\ENDWHILE
\end{algorithmic}

\end{algorithm*}

\medskip

\end{document}